\pdfoutput=1

\documentclass[11pt,a4paper]{article}

\usepackage[hyperref]{acl2021}

\usepackage{times}
\usepackage{latexsym}

\usepackage{xurl}
\usepackage{times}
\usepackage{latexsym}
\usepackage{microtype}
\usepackage{graphicx}
\usepackage{amsthm,amsmath,amssymb}
\usepackage{mathrsfs}
\usepackage{graphicx}
\usepackage{multirow}
\usepackage{color,xcolor}
\usepackage{array}
\usepackage{booktabs}
\usepackage{bm}
\usepackage{soul}
\usepackage{amsmath}
\usepackage{threeparttable}
\usepackage{tablefootnote}
\usepackage[ruled, lined, linesnumbered, commentsnumbered, longend]{algorithm2e}
\usepackage{enumitem}

\usepackage{microtype}
\definecolor{amethyst}{rgb}{0.54, 0.17, 0.89}
\definecolor{mygray}{gray}{0.6}
\newcommand{\yh}[1]{\textcolor{black}{{#1}}}
\newcommand{\gab}[1]{\textcolor{black}{{#1}}}
\newcommand{\gl}[1]{\textcolor{black}{{#1}}}
\newcommand{\hq}[1]{\textcolor{black}{{#1}}}

\usepackage{mathtools, nccmath}
\DeclarePairedDelimiter{\nint}\lfloor\rceil

\aclfinalcopy 
\def\aclpaperid{611} 

\title{Position Bias Mitigation: A Knowledge-Aware Graph Model for Emotion Cause Extraction}

\author{Hanqi Yan, Lin Gui, Gabriele Pergola, Yulan He \\
  Department of Computer Science, University of Warwick \\
  \texttt{\{hanqi.yan, lin.gui, gabriele.pergola, yulan.he\}@warwick.ac.uk} \\}

\begin{document}
\maketitle
\begin{abstract}
The \gl{Emotion Cause Extraction (ECE)} task aims to identify clauses \yh{which} \gl{contain} \yh{emotion-evoking information for a particular emotion expressed in text. We observe that a widely-used ECE dataset exhibits a bias that the majority of annotated cause clauses are either directly before their associated emotion clauses or are the emotion clauses themselves. Existing models for ECE tend to explore such relative position information and suffer from the dataset bias. To investigate the degree of reliance of existing ECE models on clause relative positions, we propose a novel strategy to generate adversarial examples in which the relative position information is no longer the indicative feature of cause clauses. We test the performance of existing models on such adversarial examples and observe a significant performance drop. To address the dataset bias, we propose a novel graph-based method to explicitly model the emotion triggering paths by } 
\gl{leveraging the commonsense knowledge to enhance the semantic dependencies between a candidate clause and an emotion clause}. 
Experimental results show that our proposed approach performs on par with the existing state-of-the-art methods on the original ECE dataset, and is more robust 
against adversarial attacks
compared to existing models.\footnote{Our code can be accessed at \url{https://github.com/hanqi-qi/Position-Bias-Mitigation-in-Emotion-Cause-Analysis}}

\end{abstract}

\section{Introduction}
Instead of detecting sentiment polarity from text, recent years have seen a surge of research activities that \yh{identify} 
the cause of emotions expressed 
in text~\citep{Gui17,article_dataset,DBLP:journals/corr/abs-1805-06533,DBLP:journals/corr/abs-1906-01267,kim-klinger-2018-feels,bostan-klinger2020}. 
In a typical dataset \yh{for} \emph{Emotion Cause Extract}~(ECE)~\citep{Gui17}, 
a document consists of multiple clauses, \yh{one of which is the emotion clause annotated with} 
a pre-defined emotion \yh{class label}. \yh{In addition, one or more clauses are annotated as the cause clause(s) which expresses triggering factors leading to the emotion expressed in the emotion clause. An emotion extraction} model \yh{trained on the dataset} is expected to classify a given clause as a cause clause or not, 
given the emotion clause. 
\begin{figure}[htb]
    \includegraphics[width=\linewidth,trim={5 5 5 5},clip]{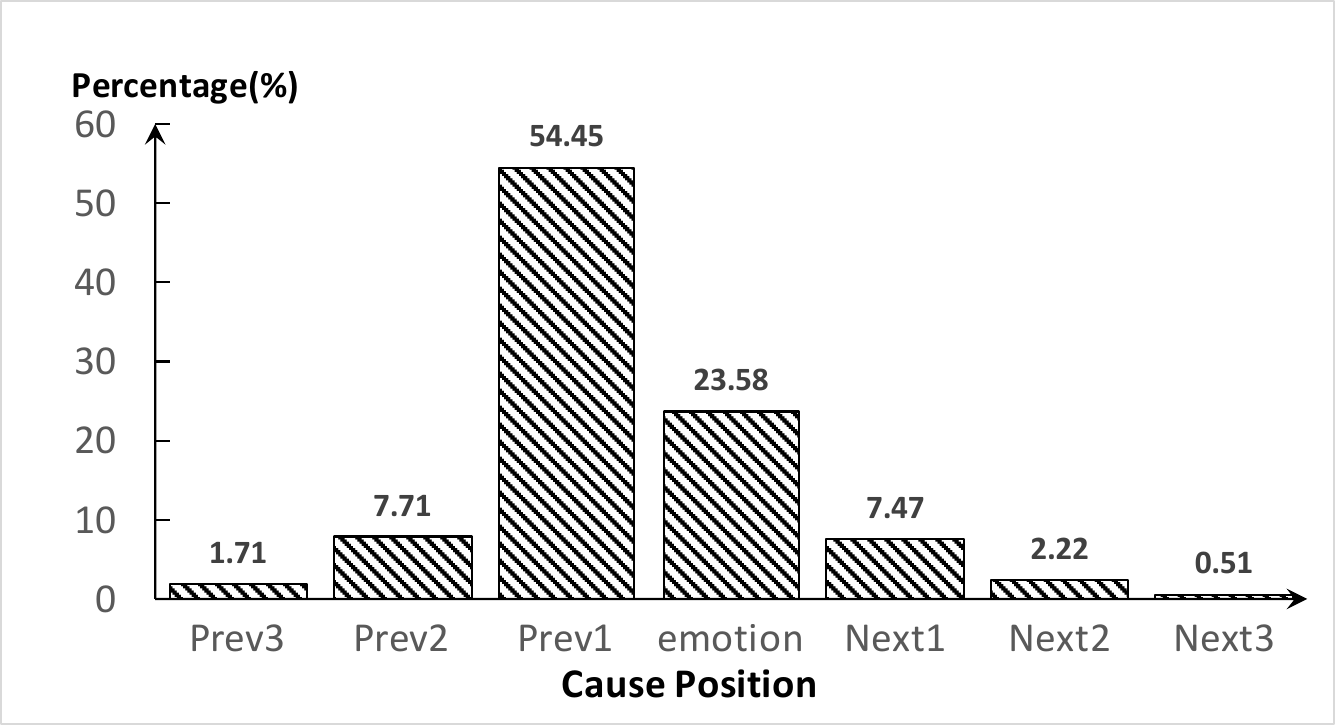}
    \caption{The distribution of positions of cause clauses relative to their corresponding emotion clauses in the ECE dataset~\citep{Gui16}. Nearly 87\% of cause clauses are located near the emotion clause (About 55\% are immediately preceding the emotion clause, 24\% are the emotion clauses themselves and over 7\% are immediately after the emotion clause).} 
    \label{Fig: data_pos_sta.}
\end{figure}


\gl{However, due to the difficulty in data collection, the ECE datasets were typically constructed by using emotion words as queries to retrieve relevant contexts as candidates for emotion cause annotation, which might lead to a strong positional bias~\cite{ding2020experimental}. Figure \ref{Fig: data_pos_sta.} depicts the distribution of positions of cause clauses relative to the emotion clause in the ECE dataset \cite{Gui16}. Most cause clauses are either immediately preceding their corresponding emotion clauses or are the emotion clauses themselves.} 
Existing ECE models tend to \gab{exploit} such relative position information and \gab{have achieved good results on emotion cause detection}.
\gl{For example,} 
The Relative Position Augmented with Dynamic Global Labels (PAE-DGL)~\cite{DingHZX19}, RNN-Transformer Hierarchical Network (RTHN)~\cite{Xia19} and Multi-Attention-based Neural Network (MANN)~\cite{Li19} all concatenate the relative position embeddings with clause semantic embeddings as the clause representations.


\yh{We argue that models utilising clause relative positions would inherently suffer from the dataset bias, and therefore may not generalise well to unseen data when} \gl{the} \yh{cause clause is not in proximity to the emotion clause. For example, in a recently released emotion cause dataset, only 25-27\% cause clauses are located immediately before the emotion clause \cite{poria2020recognizing}. 
To investigate the degree of reliance of existing ECE models on clause relative positions, we propose a novel strategy to generate adversarial examples in which the relative position information is no longer the indicative feature of cause clauses. We test the performance of existing models on such adversarial examples and observe a significant performance drop.} 

\yh{To alleviate the position bias problem,} we propose to leverage the commonsense knowledge to enhance the 
\yh{semantic} dependencies between a candidate clause and the emotion clause.
\yh{More concretely,} 
we build a clause graph, whose node features are initialised by the clause representations, \gab{and has} two types of edges~i.e., Sequence-Edge~(\textsl{S-Edge}) and Knowledge-Edge~(\textsl{K-Edge}). 
A \textsl{S-Edge} links two consecutive clauses to \gab{capture} the clause \yh{neighbourhood information}, while a \textsl{K-Edge} links a candidate clause with the emotion clause \yh{if there} \gl{exists} \gl{a knowledge path extracted from the ConceptNet \cite{DBLP:conf/aaai/SpeerCH17} between them}. 
We extend Relation-GCNs~\citep{Schlichtkrull2018Modeling} to update the graph nodes by gathering information encoded in the two types of edges. \yh{Finally,} \gl{the} \yh{cause clause is detected by performing node (i.e., clause) classification on the clause graph.} 
\yh{In summary, our contributions are three-fold:}
\begin{itemize}[noitemsep,topsep=0pt]
    \item \yh{We investigate the bias in the Emotion Cause Extraction (ECE) dataset and propose a novel strategy to generate adversarial examples in which the position of a candidate clause relative to the emotion clause is no longer the indicative feature for cause extraction.}
    \item \yh{We develop a new emotion cause extraction approach built on clause graphs in which nodes are clauses and edges linking two nodes capture the neighbourhood information as well as the implicit reasoning paths extracted from a commonsense knowledge base between clauses.} 
    \yh{Node representations are updated using the extended Relation-GCN.}
    \item \yh{Experimental results show that our proposed approach performs on par with the existing state-of-the-art methods on the original ECE dataset,} \gl{and is more robust when evaluating on the adversarial examples.} 
\end{itemize}


\begin{figure*}[t]
     \centering
     \includegraphics[width=0.95\textwidth,trim={200 250 150 60},clip]{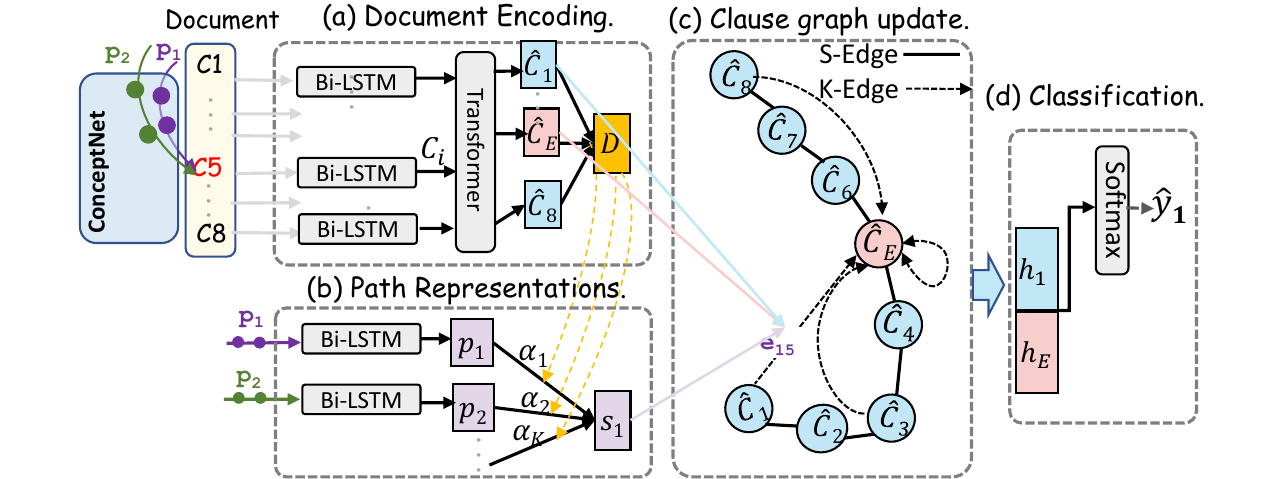}
     \caption{The framework of our proposed KAG. Given an input document consisting of eight clauses ($C1\cdots C8$), 
     we first extract knowledge paths from ConceptNet between each candidate clause and the emotion clause (\textsection \ref{sec:knowledgePath}), e.g., two knowledge paths, $p_{1}$ and $p_{2}$, are extracted between $C_{1}$ and the emotion clause $C_{5}$. \textbf{(a) Document Encoding.} Clauses are fed into a word-level Bi-LSTM and a clause-level Transformer to obtain the clause representations $\hat{\bm{C}}_{i}$. The document embedding $\bm{D}$ is generated by Dot-Attention between the emotion embedding $\bm{\hat{C}}_{E}$ and clause embeddings. \textbf{(b) Path Representations.} The extracted knowledge paths are fed into Bi-LSTM to derive path representations. Multiple paths between a clause pair are aggregated into $\bm{s}_{i}$ based on their attention to the document representation $\bm{D}$. \textbf{(c) Clause Graph Update}. A clause graph is built with the clause representations $\hat{C_{i}}$ used to initialise the graph nodes. The \textsl{K-Edge} weight $e_{iE}$ between a candidate clause $\hat{\bm{C}}_{i}$ and the emotion clause $\hat{\bm{C}}_{E}$ are measured by their distance along their path $\bm{s}_{i}$. \textbf{(d) Classification.} Node representation $\mathbf{h}_i$ of a candidate clause $C_i$ is concatenated with the emotion node representation $\mathbf{h}_E$, and then fed to a softmax layer to yield the clause classification result $\hat{\mathbf{y}}_i$.}
     \label{fig:model_overview}
 \end{figure*}

\section{Related Work}
\label{sec:related_work}

\gab{The presented work is closely related to two lines of research in emotion cause extraction: position-insensitive and position-aware models.}

\noindent \textbf{Position-insensitive Models.} A more traditional line of research exploited structural representations of textual units relying on rule-based systems \cite{Lee10} or incorporated commonsense knowledge bases \cite{Gao15} for emotion cause extraction. Machine learning  methods leveraged text features~\citep{Gui17} and combined them with multi-kernel Support Vector Machine (SVM)~\citep{Xu17}. More recent works developed neural architectures to generate effective semantic features. \citet{Cheng17} employed LSTM models, \citet{Gui17} made use of memory networks, 
while \citet{Xiangju2018A} devised a Convolutional Neural Network (CNN) with \gl{a} co-attention mechanism.
~\citep{chen-etal-2018-joint} used the emotion classification task to enhance cause extraction results.

\noindent \textbf{Position-aware Models.} More recent methodologies \gl{have} started to explicitly leverage the positions of cause clauses with respect to the emotion clause. A common strategy is to concatenate the clause relative position embedding with the candidate clause representation~\citep{DingHZX19, Xia19, Li19}. The Relative Position Augmented with Dynamic Global Labels (PAE-DGL)~\citep{DingHZX19} reordered clauses based on their distances from the target emotion clause, and propagated the information of surrounding clauses to the others. 
~\citet{8625499} used emotion dependent and independent features to rank clauses and identify the cause.
The RNN-Transformer Hierarchical Network (RTHN) \citep{Xia19} argued there exist relations between clauses in a document and proposed to classify multiple clauses simultaneously. 
\citet{Li19} proposed a Multi-Attention-based Neural Network (MANN) to model the interactions between a candidate clause and the emotion clause. The generated representations \gl{are fed} to a CNN layer for emotion cause extraction. The Hierarchical Neural Network~\cite{fan-etal-2019-knowledge} aimed at narrowing the gap between the prediction distribution $p$ and the true distribution of the cause clause relative positions.

\section{Knowledge-Aware Graph (KAG) Model for Emotion Cause Extraction}

We first define the Emotion Cause Extraction (ECE) task here. A document $\mathcal{D}$ contains $N$ clauses $\mathcal{D} = \{C_{i}\}_{i=1}^{N}$, one of which is annotated as an emotion clause $C_{E}$ with a pre-defined emotion class label, $E_{w}$.
The ECE task is to identify one or more cause clauses, $C_{t}$, $1\le t\le N$, that trigger the emotion expressed in $C_{E}$. Note that the emotion clause itself can be a cause clause.

We propose a Knowledge-Aware Graph (KAG) model as shown in Figure~\ref{fig:model_overview}, which incorporates knowledge paths extracted from ConceptNet for emotion cause extraction. \yh{More concretely, for each document, a graph is first constructed by representing each clause in the document as a node. The edge linking two nodes captures the sequential relation between neighbouring clauses (called the \emph{Sequence Edge} or \emph{S-Edge}). In addition, to better capture the semantic relation between a candidate clause and the emotion clause, we identify keywords in the candidate clause which can reach the annotated emotion class label by following the knowledge paths in the ConceptNet. The extracted knowledge paths from ConceptNet are used to enrich the \gl{relationship} between the candidate clause and the emotion clause and are inserted into the clause graph as the \emph{Knowledge Edge} or \emph{K-Edge}. We argue that by adding the \textsl{K-Edges}, we can better model the semantic relations between a candidate clause and the emotion clause, regardless of their relative positional distance.}

\begin{figure*}[t]
    \centering
    \includegraphics[width=0.9\textwidth]{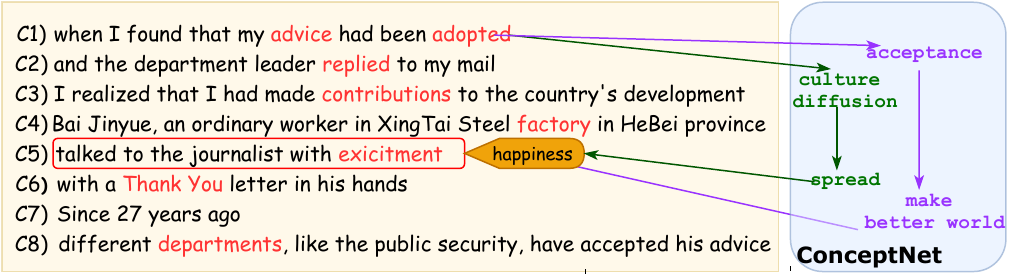}
    \caption{A document consisting of 8 clauses in the ECE dataset with extracted knowledge paths from the ConceptNet. Words in red are identified keywords. `\textit{happiness}' is the emotion label of the emotion clause $C{5}$. For better visualization, we only display two extracted knowledge paths between `\textit{adopt}' and `\textit{happiness}' in the ConceptNet.}
    \label{fig:intro_ex}
\end{figure*}


\yh{In what follows, we will first describe how to extract knowledge paths from ConceptNet, then present the incorporation of the knowledge paths into context modelling, and finally discuss the use of Graphical Convolutional Network (GCN) for learning node (or clause) representations and the prediction of the cause clause based on the learned node representations}.


\subsection{Knowledge Path Extraction from ConceptNet}
\label{sec:knowledgePath}

ConceptNet is a commonsense \yh{knowledge} 
graph, which \yh{represents entities as nodes and \gl{relationship} between them as edges}. 
To explore the causal relation between a candidate clause and the emotion clause, we \yh{propose to extract cause-related paths linking a word in the candidate clause with the annotated emotion word or the emotion class label, $E_{w}$, in the emotion clause}. 
\yh{More concretely,}
for a candidate clause, we first \yh{perform word segmentation} using the Chinese segmentation tool, Jieba\footnote{https://github.com/fxsjy/jieba}, \yh{and then} extract the top three keywords \yh{ranked by Text-Rank\footnote{We have also experimented with other keyword extraction strategies, such as extracting words with higher TFIDF values or keeping all words after removing the stop words. But we did not observe improved emotion cause detection results.}.}
Based on the findings in ~\citep{fan-etal-2019-knowledge} that sentiment descriptions can be relevant to the emotion cause, we also include adjectives in the keywords set. 

We regard each keyword in a candidate clause as a \textit{head entity}, $e_{h}$, \yh{and the emotion word or the emotion class label} in the emotion clause as the \textit{tail entity}, $e_{t}$. 
Similar to ~\citep{lin2019kagnet}, we apply networkx\footnote{http://networkx.github.io/} to perform \gl{a depth-first} search \yh{on the ConceptNet to identify \gl{the} paths which start from $e_{h}$ and end at $e_{t}$}, and only keep the paths \yh{which contain less than} 
two intermediate entities. This is because shorter paths are more likely to offer reliable reasoning evidence~\citep{wenhan_emnlp2017}. \yh{Since not all relations in ConceptNet are related to or indicative of causal relations, we further remove \gl{the} paths which contain any of these} 
four relations:~`\textsl{antonym}',~`\textsl{distinct from}',~`\textsl{not desires}', and~`\textsl{not capable of}'. Finally, we \yh{order paths by their lengths in an ascending order and} choose the top $K$ paths as the result for each candidate-emotion clause pair\footnote{We set $K$ to 15, which is the median of the number of paths between all the candidate-emotion clause pairs in our dataset.}. 

\yh{An example is shown in Figure~\ref{fig:intro_ex}. The 5-th clause is annotated as the emotion clause and the emotion class label is `\emph{happiness}'. For the keyword, `\emph{adopted}', in the first clause, we show two example paths extracted from ConceptNet, each of which links the word `\emph{adopted}' with `\emph{happiness}'. \gl{One such a path} is ``\emph{adopted} $-$\texttt{related\_to}$\rightarrow$ \emph{acceptance} $-$\texttt{has\_subevent}$\rightarrow$ \emph{make better world} $-$\texttt{causes}$\rightarrow$ \emph{happiness}''.}



\subsection{Knowledge-Aware Graph (KAG) Model}

\gl{As shown in \hq{Figure~\ref{fig:model_overview}}, there are four components in our model: a document encoding module, a context-aware path representation learning module, a GCN-based graph representation updating module, and finally a softmax layer for cause clause classification.} 

\paragraph{Initial Clause/Document Representation Learning}
\label{sec:contextModeling}

\yh{For each clause $C_i$, we derive its representation, $\bm{C}_{i}$, by using a Bi-LSTM operating on its constituent word vectors, where each word vector} 
$\bm{w}_{i} \in \mathbb{R}^{d}$ is obtained via an embedding layer. 
To capture the \yh{sequential relationship (\textsl{S-Edges}) between neighbouring clauses} 
in a document, we \yh{feed the clause sequence into a transformer architecture}. 
Similar to the original transformer incorporating the position embedding with the word embedding, we utilise the clause position information to enrich the clause representation. 
Here, the position embedding $\mathbf{o}_{i}$ of each clause is concatenated with its representation $\mathbf{C}_{i}$ generated by Bi-LSTM.
\begin{equation}\small
    \hat{\bm{C}_{i}} = \text{Transformer}(\bm{C}_{i} \,||\, \bm{o}_{i})
\end{equation}
\yh{We consider different ways for encoding position embeddings using either relative or} \gab{absolute} \yh{clause positions and explore their differences in the experiments section. In addition, we will also show the results without using position embeddings at all.}

\yh{Since the aim of our task} 
is to identify the cause clause given an emotion clause, we capture the dependencies between each candidate clause and the emotion clause. Therefore, in the document context modelling, \yh{we consider} the emotion clause $\hat{\bm{C}_{E}}$, \yh{generated in a similar way as $\hat{\bm{C}_{i}}$,} as the query vector, and the candidate clause representation $\hat{\bm{C}}_{i}$ as both the key and value vectors, \yh{in order to derive the document representation}, $\mathbf{D}\in \mathbb{R}^{d}$. 

\paragraph{Context-Aware Path Representation}
In Section~\ref{sec:knowledgePath}, we have chosen a maximum of $K$ paths $\{p_{t}\}_{t=1}^{K}$ linking each candidate $C_{i}$ with the emotion clause. However, not every path correlates equally to the document context. 
\yh{Taking the document shown} 
in Figure~\ref{fig:intro_ex} \yh{as an example, the purple knowledge path is more closely related to the document context compared to the green path. As such, we should assign \gl{a} higher weight to the purple path than the green one}.  
\yh{We propose to use the} document-level representation \gab{$\bm{D}$} \yh{obtained above as the query vector, and a knowledge path as both key and value vectors, in order to calculate the similarity between the knowledge path and the document context}. 
\yh{For each pair of a candidate clause~$C_{i}$ and the emotion clause,} we then aggregate the $K$ knowledge paths to derive the context-aware path representation $\bm{s}_{i}\in \mathbb{R}^{d}$ below: 
\begin{equation}\small
    \bm{s}_{i} = \sum_{t=1}^{K}{\alpha_{t}\bm{p}_{t}} \quad
    \alpha_{t} =  \text{softmax}(\frac{\bm{D}^{T}\bm{p}_{t}}{\sum_{j=1}^{K}{ \bm{D}^{T}\bm{p}_{j} }}) 
\end{equation}
where $\bm{D}$ is the document representation, $\bm{p}_{t}$ is the path representation obtained from Bi-LSTM on a path expressed as an entity-relation word sequence.

\paragraph{Update of Clause Representations by GCN}
\hq{After constructing a clause graph such as the one shown in Figure~\ref{fig:model_overview}(c), we update the clause/node representations via \textsl{S-Edges} and \textsl{K-Edges}. Only clauses with valid knowledge paths to the emotion clause are connected with the emotion clause node.}

\yh{After initialising the node (or clause) in the clause graph with $\hat{\bm{C}}_{i}$ and the extracted knowledge path with $\bm{s}_{i}$, we update clause representation using} 
an extended version of GCN, i.e. Relation-GCNs~(aka. R-GCNs)~\citep{Schlichtkrull2018Modeling}, which is designed for information aggregation over multiple different edges: 
\begin{equation}\small
     \bm{h}_{i}^{\ell+1} = 
     \sigma (\sum_{r\in \mathcal{R}_{N_{i}}}\sum_{j\in{N_{i}}}\frac{1}{c_{i,r}}\bm{W}_{r}^{\ell}\bm{h}_{j}^{\ell}+
     \bm{W}_{0}^{\ell}\bm{h}_{i}^{\ell})
\label{eq:r-gcns}
\end{equation}
where $\bm{W}_{r}^{\ell}\bm{h}_{j}^{\ell}$ is the linear transformed information from the neighbouring node $j$ with relation $r$ at the $\ell$-th layer, $\bm{W}_{r}^{\ell}\in \mathbb{R}^{d\times d}$ is relation-specific, $\mathcal{N}_{i}$ is the set of neighbouring nodes of the $i\text{-th}$ node, $\mathcal{R}_{N_{j}}$ is the set of distinct edges linking the current node and its neighbouring nodes.

\hq{When aggregating the neighbouring nodes information along the \textsl{K-Edge}, we leverage the path representation $\bm{s}_{i}$ to measure the node importance.} This idea is inspired by the translation-based models in graph embedding methods~\citep{NIPS2013_5071}.
Here, if a clause pair contains a possible reasoning process described by the~\textsl{K-Edge}, then $\hat{\bm{h}}_{E}\approx \hat{\bm{h}}_{i}+\bm{s}_{i}$ holds.
Otherwise, $\hat{\bm{h}}_{i}+\bm{s}_{i}$ should be far away from the emotion clause representation $\hat{\bm{h}}_{E}$.\footnote{Here, we do not consider the cases when the candidate clause is the emotion clause (i.e., $\bm{\hat{h}}_{i}=\hat{\bm{h}}_{E}$), as the similarity between $\hat{\bm{h}}_{E}+\bm{s}_{i}$ and $\hat{\bm{h}}_{E}$ will be much larger than the other pairs.} Therefore, we measure the importance of graph nodes according to the similarity between $(\bm{h}_{i}+\bm{s}_{i})$ and $\bm{h}_{E}$.
Here, we use the scaled Dot-Attention to calculate the similarity~$e_{iE}$ and obtain the updated node representation $\bm{z}_{i}$.
\begin{equation}\small
\bm{z}_{i} = \text{softmax}(\bm{e}_{E})\bm{h}^{\ell}_{E}\quad
  e_{iE} =\frac{{(\bm{h}_{i}+\bm{s}_{i})}^{T}\bm{h}_{E}}{\sqrt{d}} (i \ne E) \label{eq:eik}
\end{equation}
where $\bm{e}_{E}$ is $\{e_{iE}\}_{i=1}^{N-1}$. $d$ is the dimension of graph node representations, and $ \mathcal{N}^{r_{k}}$ is a set of neighbours by the \textsl{K-Edge}.

Then, we combine the information encoded in \textsl{S-Edge} with $\bm{z}_{i}$ as in Eq.~\ref{eq:r-gcns}, and perform a non-linear transformation to update the graph node representation $\bm{h}_{i}^{\ell+1}$:
\begin{equation}
    \bm{h}^{\ell+1}_{i} = \sigma\big(\bm{z}_{i}^{\ell} + \sum_{j \in N^{r_{s}}_{i}} (\bm{W}_{j}\bm{h}_{j})\big)
\end{equation}
where $ N_{i}^{r_{s}}$ is a set of $i$-th neighbours connected by the \textsl{S-Edges}.

\paragraph{Cause Clause Detection}
Finally, we concatenate the candidate clause node~$h_{i}$ and the emotion node representation~$h_{e}$ generated by the graph, and apply a softmax function to yield the predictive class distribution $\hat{y}_{i}$.
\begin{equation}
     \hat{y}_{i} = \mathrm{softmax}\big(\bm{W}(\bm{h}_{i}^{L} \,||\, \bm{h}_{E}^{L})+b\big),
\end{equation}

\section{Experiments}
\label{sc:exp}
We conduct a thorough experimental assessment of the proposed approach against several state-of-the-art models\footnote{Training and hyper-parameter details can be found in Appendix A.}. 

\paragraph{Dataset and Evaluation Metrics}
The evaluation dataset~\citep{Gui16} consists of 2,105 documents from SINA city news. As the dataset size is not large, we 
perform 10-fold cross-validation and report results on three standard metrics, i.e. Precision (P), Recall (R), and F1-Measure, all evaluated at the clause level.

\paragraph{Baselines}
\label{ss:baselines}
We compare our model with the position-insensitive and position-aware baselines:\\
\textbf{RB}~\cite{Lee10} and \textbf{EMOCause}~\cite{emocause:inproceedings} are rules-based methods. \textbf{Multi-Kernel}~\cite{Gui16} and \textbf{{Ngrams+SVM}}~\cite{Xu17} leverage \textbf{S}upport \textbf{V}ector \textbf{M}achines via different textual feature to train emotion cause classifiers. \textbf{CNN}~\cite{Kim2014Convolutional} and~\textbf{CANN}~\cite{Xiangju2018A} are vanilla or attention-enhanced approaches. \textbf{Memnet}~\cite{Gui17} uses a deep memory network to re-frame ECE as a question-answering task. Position-aware models use the relative position embedding to enhance the semantic features. \textbf{HCS}~\cite{yu2019multiple} uses separate hierarchical and attention module to obtain context and information. Besides that,~\textbf{PAE-DGL}~\cite{DingHZX19} and ~\textbf{RTHN} \cite{Xia19} use similar \textbf{G}lobal \textbf{P}rediction \textbf{E}mbedding~(\textit{GPE}) to twist the clauses' first-round predictions. \textbf{MANN}~\cite{Li19} performs multi-head attention in CNN to jointly encode the emotion and candidate clauses. \textbf{LambdaMART}~\citep{8625499} uses the relative position, word-embedding similarity and topic similarity as emotion-related feature to extract cause.

\begin{table}[tb]
\centering
     \resizebox{0.98\columnwidth}{!}{%
     \begin{tabular}{ll|ccc}
   \toprule[1pt]
&\textbf{Methods}& \textbf{P (\%)} & \textbf{R (\%)} & \textbf{F1 (\%)}\\ \midrule
\multirow{7}{*}{\textbf{W/O Pos}}& 
RB & 67.47 & 42.87 & 52.43 \\ 
&EMOCause  & 26.72 & 71.30 &38.87 \\ 
&Ngrams+SVM & 42.00 & 43.75 &42.85 \\
&Multi-Kernel & 65.88 & 69.27 & 67.52  \\
&CNN & 62.15 & 59.44 &60.76 \\ 
&CANN  & 77.21 & 68.91 &72.66 \\ 
&Memnet & 70.76& 68.38 &69.55 \\
   \midrule
\multirow{5}{*}{\textbf{W. Pos}}&  HCS & 73.88 & 71.54 &72.69  \\
&MANN & 78.43 & 75.87 &77.06  \\
&LambdaMART &77.20&74.99&76.08\\
&PAE-DGL & 76.19 & 69.08 &72.42\\
&RTHN & 76.97 & \textbf{76.62} & 76.77 \\
 \midrule 
 \multirow{4}{*}{\textbf{Our}}&
KAG & \textbf{79.12} &75.81&\textbf{77.43} \\
&: w/o R-GCNs&73.68& 72.76 & 73.14 \\ 
&: w/o K-Edge&75.67 & 72.63 &74.12 \\ 
&: w/o S-Edge&76.34& 75.46 & 75.88\\
   \bottomrule[1pt]
     \end{tabular}
     }
\caption{Results of different models on the ECE dataset. Our model achieves the best Precision and F1 score.}
\vspace{-5pt}
\label{tab:main_results}
\end{table}

\subsection{Main Results}
\yh{Table~\ref{tab:main_results} shows the cause clause classification results on the ECE dataset.} Two rule-based methods have poor performances, possibly due to their pre-defined rules. Multi-Kernel performs better than the vanilla SVM, \gab{being able to leverage more contextual information.} 
\gab{Across} the other three groups, the precision scores are higher than recall scores, \gab{and it is probably due to the unbalanced number of cause clauses (18.36\%) and non-cause clauses~(81.64\%), leading the models} to predict a clause as non-cause more often. 

Models in the position-\gab{aware} group perform better than those in the other groups, indicating the importance of position information. 
Our proposed model \yh{outperforms all the other models except RHNN in which its recall score is slightly lower. We have also performed ablation studies by removing either \textsl{K-Edge} or \textsl{S-Edge}, or both of them (w/o \textsl{R-GCNs}). The results show that} 
removing the R-GCNs leads to a drop of nearly 4.3\% in F1. Also, both the \textsl{K-Edge} and \textsl{S-Edge} contributes to emotion cause extraction. As contextual modelling has considered the position information, the removal of \textsl{S-Edge} leads to a smaller drop compared to the removal of \textsl{K-Edge}.


\subsection{Impact of Encoding Clause Position Information}

\begin{figure}[tb]
    \centering
    \includegraphics[width=\linewidth, trim=14 24 34 25, clip]{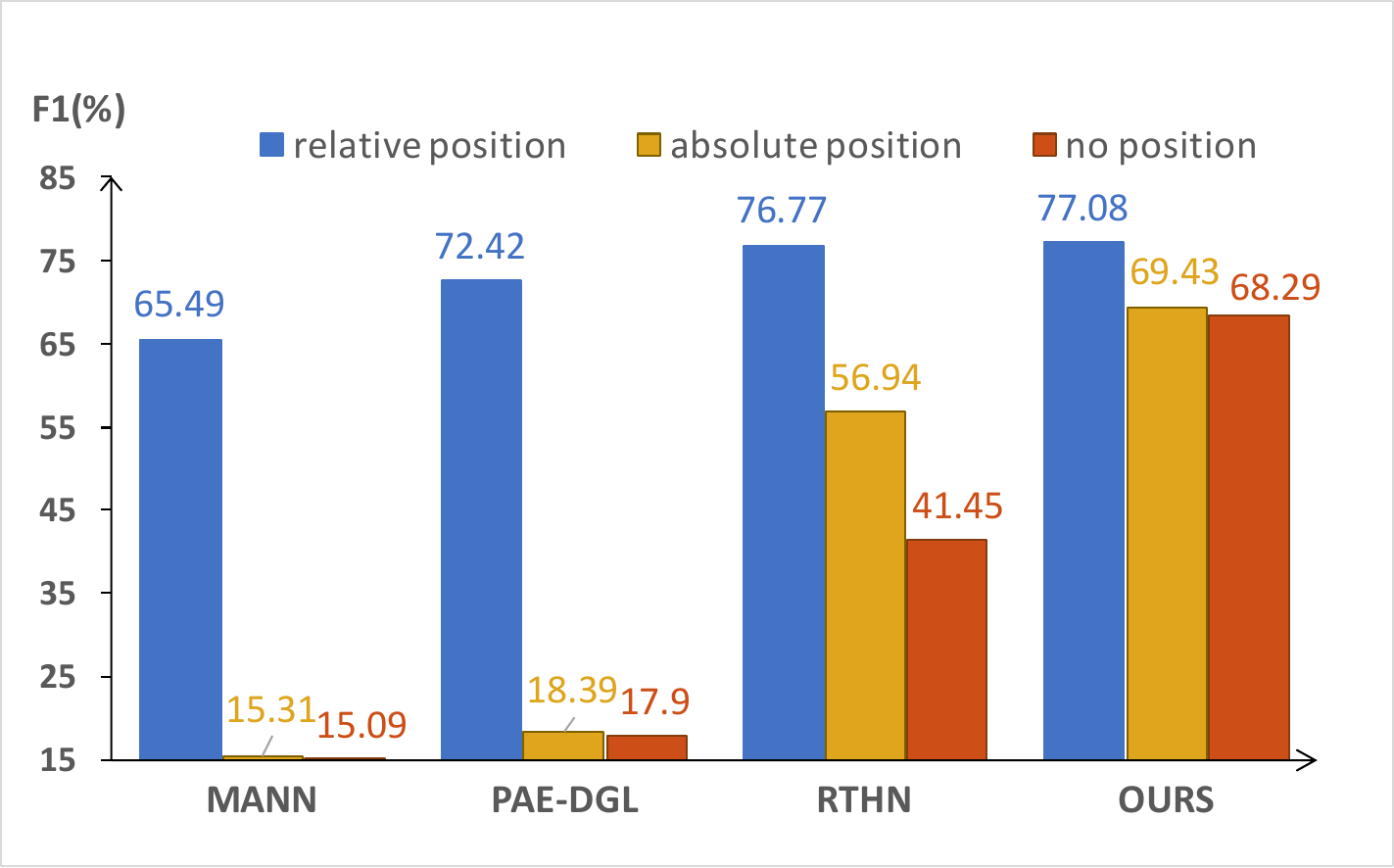}
    \caption{Emotion cause extraction when using relative, absolute or no clause positional information. Our model demonstrates most stable performance without the relative position information.}
\label{fig:absolute_pos}
\end{figure}

In order to examine the impact of using the clause position \yh{information} in different models, we replace the relative position information \yh{of \gl{the} candidate clause} with absolute positions. \yh{In the extreme case, we} remove the position information from the models. The results are shown in Figure~\ref{fig:absolute_pos}. \yh{It can be observed that the best results are achieved using relative positions for all models. Replacing relative positions using either absolution positions or no position at all \gl{results} in a significant performance drop. In particular, MANN and PAE-DGL have over 50-54\% drop in F1. The performance degradation is less significant for RTHN, partly due to its use of the Transformer architecture for context modeling. Nevertheless, we have observed a decrease in F1 score in the range of 20-35\%. } 
Our proposed model is less sensitive to the relative positions \yh{of candidate clauses}. 
Its robust performance \yh{partly attributes to the use of} 
(1) hierarchical contextual modeling via the Transformer structure, and (2) the 
\textsl{K-Egde} which helps explore causal links via commonsense knowledge regardless of a clause's relative position. 

\subsection{Performance under Adversarial Samples}
\hq{In recent years, there have been growing interests in understanding vulnerabilities of NLP systems~\cite{goodfellow2015explaining,ebrahimi2017hotflip,wallace2019universal,jin2020bert}. Adversarial examples explore regions where the model performs poorly, which could help understanding and improving the model. Our purpose here is to evaluate if KAG is vulnerable as existing ECE models when the cause clauses are not in proximity to the emotion clause.
Therefore, we propose a principled way to generate adversarial samples such that the relative position is no longer an indicative feature for the ECE task.}

\paragraph{Generation of adversarial examples} 
\hq{We generate adversarial examples to trick ECE models, which relies on swapping two clauses $C_{r_{1}}$ and $C_{r_{2}}$, where $r_1$ denotes the position of the most likely cause clause, while $r_2$ denotes the position of the least likely cause clause.} 

\yh{We identify $r_{1}$ by locating the most likely cause clause based on its relative position with respect to the emotion clause in a document. As illustrated in Figure~\ref{Fig: data_pos_sta.}, over half of } \gl{the cause clauses} \yh{are immediately before the emotion clause in the dataset. We assume that the position of a cause clause can be modelled by a Gaussian distribution and estimate} \gl{the} \yh{mean and variance directly from the data, which are, $\{\mu,\sigma^2\}=\{-1, 0.5445\}$. The position index $r_1$ can then be sampled from the Gaussian distribution. As the sampled value is continuous, we round the value to its nearest integer: }
\begin{equation}\small
    r_{1} \gets\nint{g}, \quad g \backsim \text{Gaussian}(\mu, \sigma^2).
    \label{eq:r1}
\end{equation}


\yh{To locate the least likely cause clause, we propose to choose the value for $r_2$ according to the attention score between a candidate clause and the emotion clause. Our intuition is that if the emotion clause has a lower score attended to a candidate clause, then it is less likely to be the cause clause.}
We use an existing emotion cause extraction model to generate contextual representations and 
use the Dot-Attention~\cite{luong15} to measure the similarity between each candidate clause and the emotion clause. 
\yh{We then select the index $i$ which gives the lowest attention score and assign it to $r_2$:} 

\begin{equation}\small
     r_{2} =\arg\min_{i} \{\lambda_i\}_{i=1}^{N}, \quad
     \mathbf{{\lambda}}_{i} = \text{Dot-Att.}(\hat{\mathbf{C}_{i}},\hat{\bm{C}_{E}}),
    \label{eq:r2}
\end{equation}
\yh{where $\hat{\mathbf{C}_{i}}$ is the representation of the $i$-th candidate clause, $\hat{\mathbf{C}_{E}}$ is the representation of the emotion clause, and $N$ denotes a total of $N$ clauses in a document.}

\hq{Here, we use existing ECE models as different discriminators to generate different adversarial samples.\footnote{The adversarial sample generation is independent from their training process.} The desirable adversarial samples will fool the discriminator to predict the inverse label.}
\hq{We use leave-one-model-out to evaluate the performance of ECE models. In particular, one model is used as a \emph{Discriminator} for generating adversarial samples which are subsequently used to evaluate the performance of other models.} 

\paragraph{Results} 
The results are shown in Table~\ref{tab:ad_results}. \hq{The attacked ECE models are merely trained on the original dataset}. The generated adversarial examples are used as the test set only.
\yh{We can observe a significant performance drop of 23-32\% for the existing ECE models, }
some of which even \gl{perform} worse than the earlier rule-based methods, \yh{showing their sensitivity to the positional bias in the dataset}. 
\yh{We also observe the performance degradation of} our proposed KAG. \yh{But its performance drop is less significant compared to other models. The results verify the effectiveness of} 
capturing the semantic dependencies \yh{between a candidate clause and the emotion clause} via contextual and commonsense knowledge encoding.

\begin{table}[t]
\centering
\resizebox{0.48\textwidth}{!}{
\begin{tabular}{p{2.3cm}|c|c|c|c}
\toprule[1pt]
\hline
\multirow{2}{*}{\textbf{Discriminator}}&\multicolumn{4}{c}{\textbf{Attacked ECE models}}\\
\cline{2-5}
&PAEDGL&MANN&RTHN&KAG\\
\hline
PAEDGL& 49.62 & 48.92 & 59.73&64.98 \\
&$\downarrow$31.76\%&$\downarrow$28.6\%&$\downarrow$ 22.20\%&$\downarrow$\textbf{16.08}\%\\
\hline
MANN&51.82& 47.24&60.13&66.32\\
&$\downarrow$28.45\% &$\downarrow$31.27\% &$\downarrow$21.65\%&$\downarrow$\textbf{14.35}\%\\
\hline
RTHN&48.63& 49.63 &57.78&63.47\\
&$\downarrow$32.85\% &$\downarrow$ 27.64\%&$\downarrow$ 24.74\%&$\downarrow$\textbf{18.03}\%\\
\hline
KAG & 48.52 & 48.24 &59.53&62.39\\
&$\downarrow$ 33.00\%& $\downarrow$29.67\%&$\downarrow$22.46\%&$\downarrow$\textbf{19.42}\%\\
\hline
Ave. Drop(\%)&$\downarrow$31.51\%&$\downarrow$29.29\%&$\downarrow$22.62\%&\textbf{$\downarrow$16.97\%}\\
 \bottomrule[1pt]
\end{tabular}
}
\caption{\hq{F1 score and relative drop (marked with $\downarrow$) of different ECE models on adversarial samples. The listed four ECE models are attacked by the adversarial samples generated from the respective discriminator. Our model shows the minimal drop rate comparing to other listed ECE models across all sets of adversarial samples.}} 
\label{tab:ad_results}
\end{table}


\subsection{Case Study and Error Analysis}
To understand how KAG aggregate information based on different paths, we randomly choose two examples to visualise the attention distributions~(Eq.~\ref{eq:eik}) on different graph nodes (i.e., clauses) in Figure~\ref{fig:attention weights}.\footnote{More cases can be found in the Appendix.} These attention weights show the `\emph{distance}' between a candidate clause and the emotion clause during the reasoning process. The cause clauses are underlined, and keywords are in bold. \yh{$C_i$ in brackets indicate the relative clause position to the emotion clause (which is denoted as $C_0$).}

\paragraph{Ex.1}{\small{The \textbf{crime} that ten people were \textbf{killed} shocked the whole country $(C_{-4})$. This was due to personal grievances ($C_{-3}$).  \ul{Qiu had \textbf{arguments} with the management staff} ($C_{-2}$), \ul{and thought the Taoist temple host had \textbf{molested} his wife} ($C_{-1}$). He became \textbf{angry} ($C_{0}$), and  killed the host and destroyed the temple~($C_{1}$).}}

In Ex.1, \yh{the emotion word is `\emph{angry}', the knowledge path identified by our model from ConceptNet is,} ``\emph{arguments} $\rightarrow$ \emph{fight} $\rightarrow$\emph{angry}'' for Clause $C_{-2}$, and ``\emph{molest} $\rightarrow$ \emph{irritate} $\rightarrow$\emph{exasperate}$\rightarrow$\emph{angry}'' for Clause $C_{-1}$. Our model assigns the same attention weight to the clauses $C_{-2}$, $C_{-1}$ and the emotion clause, as shown in Figure~\ref{fig:attention weights}. This shows that both paths are equally weighted by our model. Due to the \textsl{K-Edge} attention weights, our model can correctly identify both $C_{-2}$ and $C_{-1}$ clauses as the cause clauses. 

\paragraph{Ex.2} {\small{The LongBao Primary school locates between the two villages ($C_{-2}$). \ul{Some \textbf{unemployed} people always cut through the school to take a shortcut} ($C_{-1}$). Liu Yurong \textbf{worried} that it would affect children's study ($C_{0}$). When he did not have teaching duties ($C_{1}$), he stood \textbf{guard} outside the school gate ($C_{2}$).}} 

In Ex.2, the path identified by our model from ConceptNet for Clause ($C_{-1}$) is 
``\emph{unemployment} $\rightarrow$ \emph{situation} $\rightarrow$ \emph{trouble/danger}$\rightarrow$ \emph{worried}''. It has been assigned the largest attention weight as shown in Figure~\ref{fig:attention weights}. \yh{Note that the path identified is spurious since the emotion of `\emph{worried}' is triggered by `\emph{unemployment}' in the ConceptNet, while in the original text, `\emph{worried}' is caused by the event, `\emph{Unemployed people cut through the school}'. This shows that simply using keywords or entities searching for knowledge paths from commonsense knowledge bases may lead to spurious knowledge extracted. We will leave the extraction of event-driven commonsense knowledge as future work.} 

\begin{figure}[htb]
\centering
\includegraphics[width=0.48\textwidth,trim=2 2 2 2,clip]{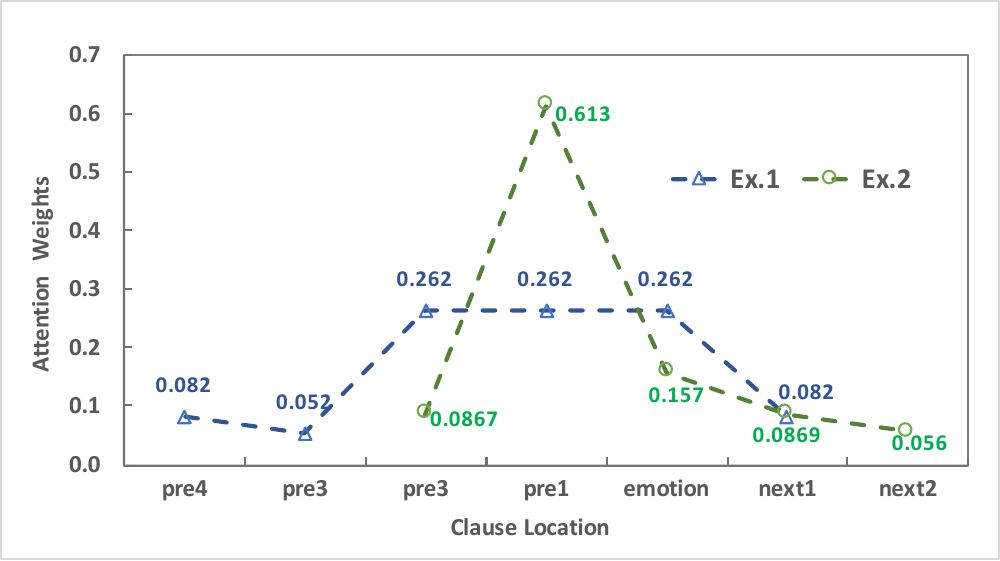}
\caption{Attention weights among different graph nodes/clauses on \textbf{Ex.1} and \textbf{Ex.2}.}
\label{fig:attention weights}
\vspace{-3mm}
\end{figure}

\section{Conclusion and Future Work}
In this paper, we \yh{examine the positional bias in the annotated ECE dataset and investigate the degree of reliance of the clause position information in existing ECE models.} 
We design a novel approach for generating adversarial samples. 
Moreover, we propose a graph-based model to enhance the semantic dependencies between a candidate clause and a given emotion clause \yh{by extracting relevant knowledge paths from ConceptNet}. 
The experimental results show that our proposed method achieves   comparative performance to the state-of-the-art methods, and \yh{is more robust against  adversarial attacks. Our current model extracts knowledge paths linking two keywords identified in two separate clauses.} 
In the future, we will exploit how to incorporate the event-level commonsense knowledge to improve the performance of emotion cause extraction.


\section*{Acknowledgements}
 This work was funded by the EPSRC (grant no. EP/T017112/1, EP/V048597/1). HY receives the PhD scholarship funded jointly by the University of Warwick and the Chinese Scholarship Council. YH is supported by a Turing AI Fellowship funded by the UK Research and Innovation (grant no. EP/V020579/1). We thank Yizhen Jia and Daoye Zhu for their valuable work on earlier code framework of this paper. We also thank the anonymous reviewers
for their valuable comments.

\bibliographystyle{acl_natbib}
\bibliography{old}
\newpage

\appendix
\section{Model Architecture}
In this section, we describe the details of the four main components in our model: \emph{contextual modelling}, \emph{knowledge path encoding}, \emph{clause graph update} and \emph{cause clause classification}. 
 
 The dataset has 2,105 documents. The maximum number of clauses in a document is 75 and the maximum number of words per clause is 45. So we first pad the input documents into a matrix $\mathbf{I}$ with the shape of $[2105,75,45]$.
 
 \subsection{Contextual Modelling}
\label{sec:contextModeling}
\underline{a. \textbf{token} $\rightarrow$\textbf{ clause}} We first apply a 1-layer Bi-LSTM of 100 hidden units to obtain word embeddings, $w\in \mathbb{R}^{200}$. 
We then use two linear transformation layers (hidden units are [200,200],[200,1]) to map the original $w$ to a scalar attention score $\alpha$, then perform a weighted aggregation to generate the clause representation $\mathbf{\hat{C}}_{i}\in \mathbb{R}^{200}$.
\\
\underline{ b. \textbf{clause} $\rightarrow$ \textbf{document}}
We feed the clause representations into a Transformer. It has 3 stacked blocks, with the multi-head number set to 5, and the dimension of \textit{key}, \textit{value}, \textit{query} is all set to 200. 
The query vector is the emotion clause representation $\mathbf{\hat{C}}_{E} \in \mathbb{R}^{200}$, the key and value representations are candidate clause representations, also with 200 dimensions. Finally, the updated clause representations are aggregated via Dot-Attention to generate the document representation $\bm{D}\in \mathbb{R}^{200}$.

\subsection{Knowledge Path Encoding}
For each candidate clause and the emotion clause, we extract knowledge paths from ConceptNet and 
only select $K$ paths. 
The values of $K$ is set to 15, since the median of the number of paths between a candidate clause and the emotion clause is 15 in our dataset.

We use the same Bi-LSTM described in Section \ref{sec:contextModeling} to encode each knowledge path and generate the K number of path representations $\{\bm{p}_{it}\}_{t=1}^{K}$ between the $i$-th clause and the emotion clause. 
Then, the document representation $\bm{D}$ is applied as the query to attend to each path in $\{\bm{p}_{it}\}$ to generate the final context-aware path representation $\bm{s}_{i}\in\mathbb{R}^{200}$. 

\subsection{Clause Graph Update}

The graph nodes are initialised by clause presentations, with the feature dimension 200. 
To calculate the attention weights $e_{iE}$ in R-GCNs, 
We use the non-linearly transformed $\bm{h}_{i}+\bm{s}_{i}$ as the query, the non-linearly transformed $\bm{h}_{E}$ as the value and key. The non-linear functions are independent Selu layers.


\subsection{Cause Clause Classification}
The MLP with [400,1]
hidden units takes the concatenation of each
candidate node $\{\bm{h}_{i}^{L}\}_{i=1}^{N}$ and the emotion node  representation $\bm{h}_{E}^{L}$ to predict the logit, after which, a softmax layer is applied to predict the probability of the cause clause.


\section{Training Details for KAG}

We randomly split the datasets into 9:1 (train/test).
For each split, we run 50 iterations to get the best model \gab{on the validation set, which takes an average time of around} 23 minutes per split, when conducted on a NVIDIA GTX 1080Ti. For each split, we test the model on the \gab{test set} at the end of each iteration and \gab{keep} the best \gab{resulting} F1 \gab{of the split}. The number of model parameters is 1,133,002.

\paragraph{Hyper-parameter Search}
We use the grid search to find the best parameters for our model on the validation data, \gab{and report in the following the hyper-parameter values providing the best performance.}
\begin{itemize}
    \item The word embeddings used to initialise the Bi-LSTM is provided by NLPCC\footnote{https://github.com/NUSTM/RTHN/tree/master/data}. It was pre-trained on a 1.1 million Chinese Weibo corpora \gab{following the Word2Vec algorithm}. The \gab{word embedding dimension} is \gab{set to} 200.
    \item The position embedding \gab{dimension} is set to 50, \gab{randomly} initialised with the uniform distribution (-0.1,0.1).
    \item The number of Transformer blocks is 2 and the number of graph layers is 3.
    \item To \gab{regularise} against over-fitting, we employ dropout (0.5 in the encoder, 0.2 in the graph layer).
    \item  The network is trained \gab{using the} the Adam optimiser with a mini-batch size 64 and a learning rate $\eta=0.005$. The parameters of our model are initialised with Glorot initialisation.
\end{itemize}


\section{Error Analysis}
We perform error analysis to \gab{identify the} \gab{limitations of the proposed model.} 
In the following examples (Ex.1 and Ex.2), the cause clauses are in bold, our predictions are underlined.


\paragraph{Ex.1}{Some kind people said ($C_{-6}$}), if Wu Xiaoli could find available kidneys ($C_{-5}$), they would like to donate for her surgery ($C_{-4}$). 4000RMB donation had been sent to Xiaoli ($C_{-3}$), Qiu Hua said ($C_{-2}$). \ul{The child's desire to survival shocked us ($C_{-1}$}). \textbf{\ul{The family's companion was touching ($C_{0}$)}}. Wish kind people will be ready to give a helping hand ($C_{1}$). Help the family in difficulty ($C_{2}$). 

In the first example~\textbf{Ex.1}, our model \gab{identifies} the keyword~\textit{\textbf{survival}} in $C_{-1}$ and extract\gab{s} several paths from `\textit{survival}' to `\textit{touching}'. However, the main event in clause~$C_{-1}$ \gab{concerns} \textit{\textbf{desire}} \gab{rather than} \textit{\textbf{survival}}. Our current model detects the emotion reasoning process from ConceptNet based on keywords identified in text, and inevitably introduces spurious knowledge paths to model learning. 

\paragraph{Ex.2}{I have only one daughter ($C_{0}$)}, and a granddaughter of 
8 year-old ($C_{-10}$). I would like to convey these memory to her ($C_{-9}$). Last Spring Festival ($C_{-8}$), I gave the DVD away to my granddaughter ($C_{-7}$). I hope she can inherit my memory ($C_{-6}$). Thus ($C_{-5}$), I feel like that my ages become eternity ($C_{-4}$). Sun Qing said ($C_{-3}$). \ul{His father is a sensitive and has great passion for his life ($C_{-2}$)}. \textbf{He did so ($C_{-1}$)}. Making me feel touched ($C_{0}$). His daughter said ($C_{1}$). 

In the \textbf{Ex}~2, our model detected the \textit{\textbf{passion}} as a keyword and extracted knowledge paths between the clause~$C_{-2}$ and the emotion clause. However, it ignores the semantic dependency between the clause~$C_{-1}$ and the emotion clause. It is therefore more desirable to consider semantic dependencies or discourse relations between clauses/sentences for emotion reasoning path extraction from external commonsense knowledge sources.

 \section{Human Evaluation on the Generated Adversarial Samples}
 The way adversarial examples generated changes the order of the original document clauses. Therefore, we would like to find out if such clause re-ordering changes the original semantic meaning and if these adversarial samples can be used to evaluate on the same emotion cause labels.
 
 We randomly selected 100 adversarial examples and ask two independent annotators to manually annotate emotion cause clauses based on the same annotation scheme of the ECE dataset. Compared to the original annotations, Annotator 1 achieved 0.954 agreement with the cohen’s kappa value of 0.79, while Annotator 2 achieved 0.938 agreement with the cohen’s kappa value of 0.72. This aligns with our intuition that an emotion expressed in text is triggered by a certain event, rather than determined by relative clause positions. A good ECE model should be able to learn a correlation between an event and its associated emotion. This also motivates our proposal of a knowledge-aware model which leverages commonsense knowledge to explicitly capture event-emotion relationships.
\end{document}